# Computational Design and Co-Robotic Fabrication for Material Reuse in Architecture

Arash Adel
Princeton University

Daniel Ruan
Princeton University

Ruxin Xie
Princeton University

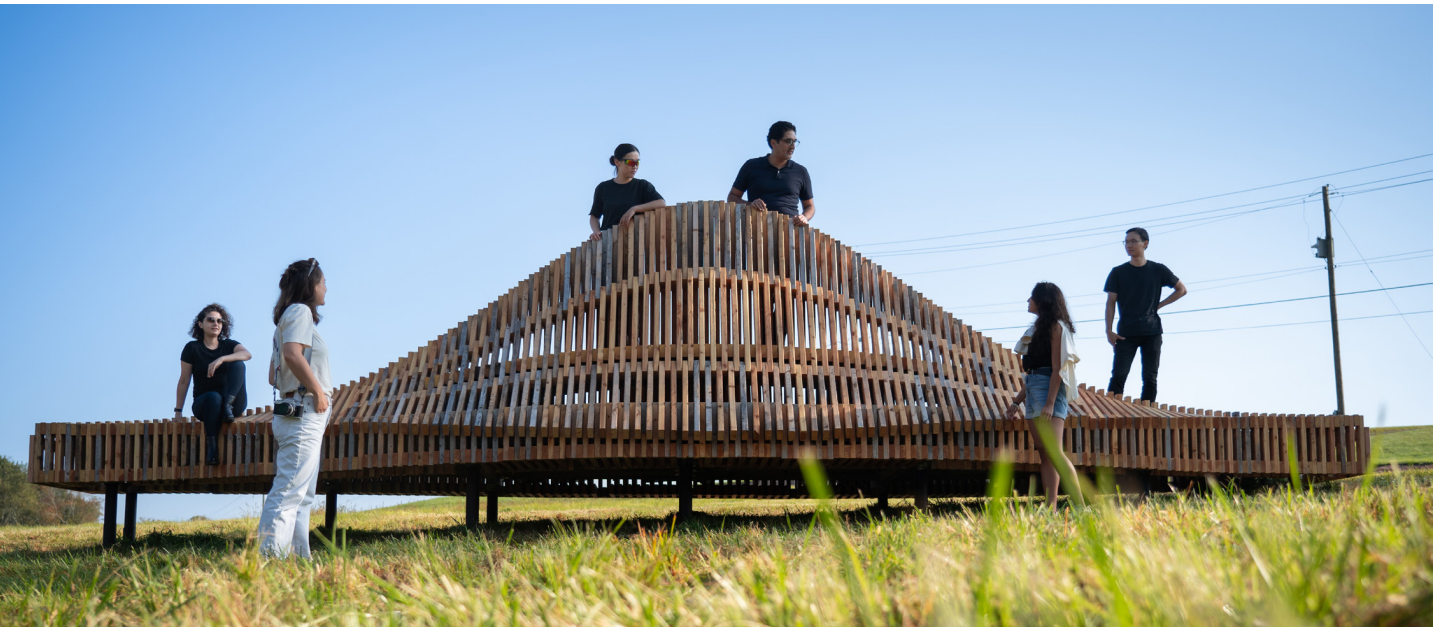

1   Timbrelyn.

ABSTRACT

Climate change and resource depletion demand a shift from the dominant linear "take-make-use-dispose" paradigm of construction toward circular, low-waste practices. Material reuse offers a promising pathway by reducing raw material extraction, mitigating waste, and extending the service lifespan of carbon-sequestering materials such as timber. Realizing this potential, however, requires addressing technical and logistical challenges across both design and construction for accommodating heterogeneous, reclaimed material inventories.

This paper presents an integrated framework that couples data-driven computational design with feedback-driven adaptive human–robot collaborative (co-robotic) fabrication and assembly to enable the realization of nonstandard structures made from reclaimed timber of varying length and geometries, supplemented with new off-the-shelf timber when necessary. The framework is validated through Timbrelyn, a built case-study installation that demonstrates how timber reuse can inform and enhance architectural expression. This work contributes to the development of integrated design-to-fabrication workflows that advance adaptive, feedback-driven methods to handle inventory constraints and reclaimed material uncertainties, facilitating material reuse in the design and construction of new buildings and structures.

INTRODUCTION

The prevailing paradigm for constructing new buildings and civil infrastructure relies heavily on extracted raw materials, amounting to roughly 2.4 billion metric tons in the United States in 2020 alone (Matos 2022). According to the United Nations (United Nations, Department of Economic and Social Affairs, Population Division 2019), by 2050, nearly 68% of the global population will reside in cities, adding approximately 2.5 billion urban residents. As demand intensifies for housing, infrastructure, and resilient urban environments, this linear "take-make-use-dispose" paradigm (Dokter, Thuvander and Rahe 2021) is increasingly unsustainable. Two critical challenges drive this shift: the need to decarbonize buildings by reducing embodied carbon in new construction and maintenance, and the growing scarcity of raw materials from excessive extraction (Ruuska and Häkkinen 2014). Addressing these challenges requires the architecture and construction sectors to rethink the dominant linear paradigm and embrace circular strategies.

One promising approach is the reclamation and reuse of building materials to support circular design and construction practices. Reusing existing components reduces demand for raw material extraction, alleviating pressure on natural resources (Dokter, Thuvander and Rahe 2021), and typically results in a lower net embodied carbon impact for new construction by avoiding emissions associated with the production of virgin materials (Tingley, Giesekam and Cooper-Searle 2018). Moreover, material reuse reduces landfill waste and extends the lifespan of carbon already sequestered in bio-based products such as timber (Grüter et al. 2023). Construction and demolition waste was estimated to be 600 million tons in 2018 in the United States (US EPA 2020), which can be substantially reduced through low-waste material reuse strategies.

This research focuses on timber as a case-study material since it is the predominant structural material used in single-family housing in the United States (94% in 2022; Fu 2023). However, timber reuse introduces technical and logistical challenges across both design and construction. To address design-related challenges, recent works in computational design have developed inventory-aware methods that treat available, often reclaimed, timber stock as a primary input. For instance, Gattas et al. (2025) presented an optimization framework for stock-constrained structural design of timber frame buildings using mixed reclaimed timber inventories. However, this and similar frameworks typically depend on labor-intensive digitization of reclaimed inventories, a process that becomes increasingly impractical for large-scale material stocks (Haakonsen et al. 2024; Bergsagel and Heisel 2023). Gattas et al. (2025) further suggested integrating digital fabrication methods to streamline the reprocessing of reclaimed timber elements. Such integration becomes essential when working with the highly heterogeneous elements typical of reclaimed inventories, where fabrication and assembly processes must accommodate sorting, labeling, and accessing elements in a specific sequence. Moreover, conventional phase-based design-to-fabrication workflows are inherently inadequate for capturing material uncertainties (e.g., dimensional tolerances) or for implementing adaptive strategies to handle these uncertainties without cascading rework.

Implementing just-in-time robotic fabrication processes offers a compelling solution to these challenges. Unlike conventional fabrication workflows, robotic assembly can accommodate material irregularity through sensing and adaptive decision-making (Adel et al. 2024; Wu and Kilian 2018). Within this context, recent studies have demonstrated growing interest in using reclaimed timber within robotic fabrication workflows. Larsen and Aagaard (2020) demonstrated robotic processing of discarded, crooked sawlogs for constructing a lamella roof structure. Cognoli, Cocco, and Ruggiero (2024) developed an upcycling workflow that combines photogrammetric surveying, building information modeling (BIM) techniques, and robotic fabrication to reuse timber from postdisaster housing. Bruun et al. (2024) presented a scaffold-free cooperative robotic system for disassembling and reassembling timber structures, emphasizing sequence planning and structural support hierarchies. Each project successfully introduced computational algorithms that adapt robotic fabrication to the geometric irregularities of scanned material inventories.

Nevertheless, because these methods depend on fully digitized material inventories, a gap remains regarding robotic fabrication methods under uncertainty of the material inventory, that is, when exhaustive inventory digitization is infeasible prior to fabrication due to time, cost, or scale constraints. Furthermore, inventory-aware computational design and just-in-time robotic assembly often remain disconnected, lacking a fully integrated design-to-fabrication workflow.

To address these gaps, this paper proposes an integrated framework that couples data-driven computational design with feedback-driven, adaptive human–robot collaborative (co-robotic) assembly to enable the realization of nonstandard structures made from reclaimed timber of varying length and geometries, supplemented with new off-the-shelf timber when necessary. The developed

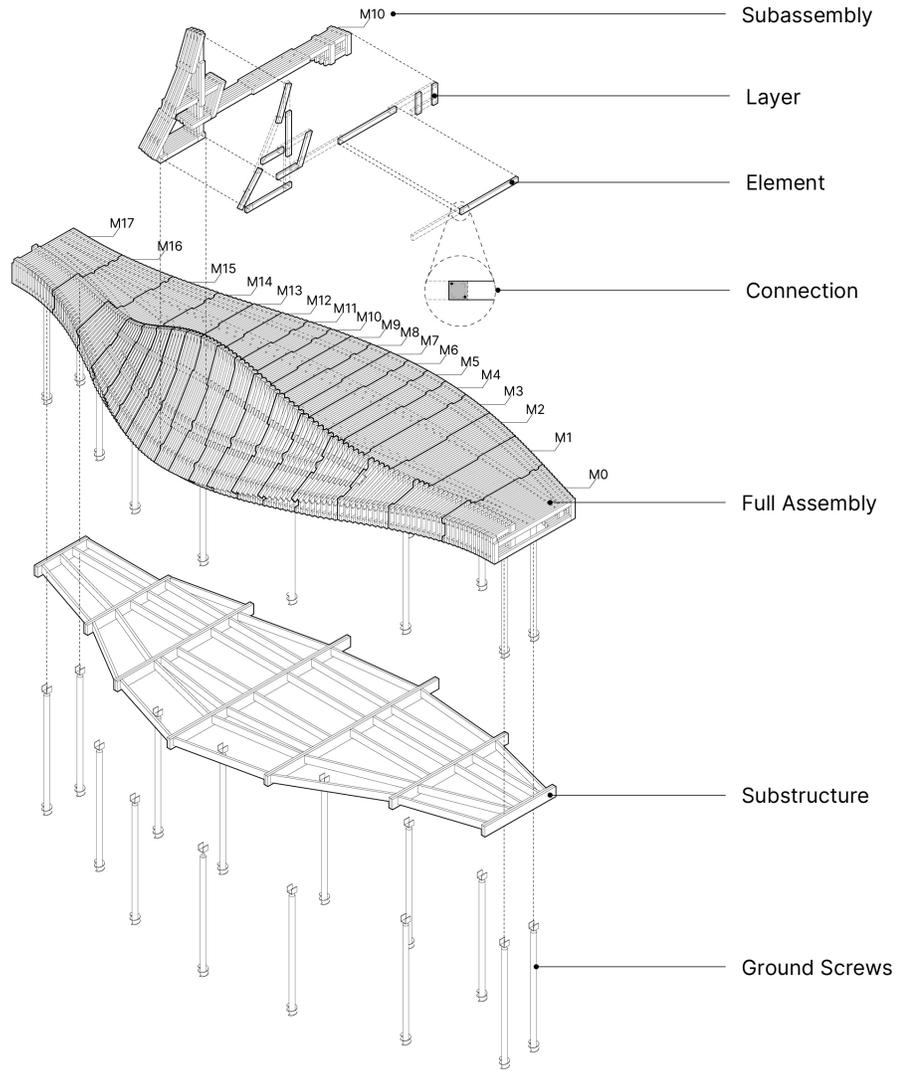

2  Exploded diagram of Timbrelyn and its components.

computational design method supports data-driven design exploration (e.g., stock lengths and quantities) while incorporating robotic fabrication, transportation, and construction logistics constraints (e.g., working envelope, subassembly weight). The developed co-robotic assembly method accommodates material irregularity (e.g., variable lengths and profiles) for fabricating bespoke subassemblies through perception-driven material selection, as-built scanning, and adaptive planning, while minimizing offcut waste. The framework is demonstrated and validated through a built case study, Timbrelyn (Fig. 1), located at the Bethel Woods Center for the Arts in Bethel, New York, the historic site of the 1969 Woodstock festival.

Nomenclature
We define several key terms used throughout our constructive system (illustrated in Fig. 2):

- Full assembly: Refers to the fully assembled installation on site, not including ground screws and substructure.

- Subassembly: Refers to a preassembled timber component, co-robotically fabricated off site.

- Layer: In the developed constructive system for the project, timber elements are arranged through a sequential stacking process to form subassemblies. Within this constructive system, each layer comprises a set of timber elements lying on a common plane.

- Element: Refers to a discrete timber piece, computationally represented by its cross-section, length, cut planes, and pose, and serves as the base unit for computational design and co-robotic fabrication.

- Connection: Refers to an overlap between two elements and its mechanical fasteners (e.g., nails or screws).

METHODS

The design, planning, and construction process of Timbrelyn followed a hybrid off-site and on-site process (Fig. 3), building on our previous research project (Adel 2022). Timber subassemblies were fabricated off-site using a co-robotic assembly method, where robots executed precise fabrication and placement tasks while human collaborators performed material preparation, quality checks, fastening, and exception handling. These subassemblies were then transported to the site, where human workers assembled them into the final installation.

More specifically, for this project, we implemented a workflow that connects disassembly and material reclamation with rough inventory classification, data-driven computational design, and adaptive co-robotic assembly (Fig. 3). This workflow begins with disassembly and material recovery, where components from our previous project, Robotically Fabricated Structure (Adel 2022), were deconstructed to preserve structural elements, and their fasteners (e.g., nails and screws) were removed. Following disassembly, we roughly sorted the reclaimed timber stock into discrete bins based on their length (e.g., 400–500 mm) to form an estimate of their quantity and distribution. This data informed the computational design method, enabling data-driven exploration of the installation design while providing metrics on material-specific constraints such as element length, cross-sectional variability, and connection strategies.

Overall, by coupling disassembly and material reclamation with an adaptive computational design-to-fabrication pipeline, the methods establish a framework for material reuse in architecture and construction. In the following sections, we elaborate on the computational design and the adaptive co-robotic assembly methods.

Computational Design

Building upon our previous research (Ruan and Adel 2023; Adel 2022), we utilize a layer-based constructive system for aggregating reclaimed lumber, augmented with newly sourced, short dimensional lumber to produce building-scale subassemblies. This system follows long-standing precedents in layered timber assemblies, more specifically, nail-laminated timber, where elements typically with a uniform cross-section are stacked in ordered strata to form larger geometries satisfying required structural integrity. In this project, reclaimed short 2×4s exhibit minimal variation in width and depth, while their lengths can vary significantly. A layer-based constructive system utilizes this asymmetry; constant cross-section simplifies registration (e.g., a planar surface for the floor), creating necessary overlaps for connections and ease of fastening, while variable lengths become a design driver for patterning and modulation of the elements across the full assembly.

A key feature of the constructive system is the use of side-grain lap connections between adjacent layers (Fig. 2). Because the side faces of elements in neighboring layers

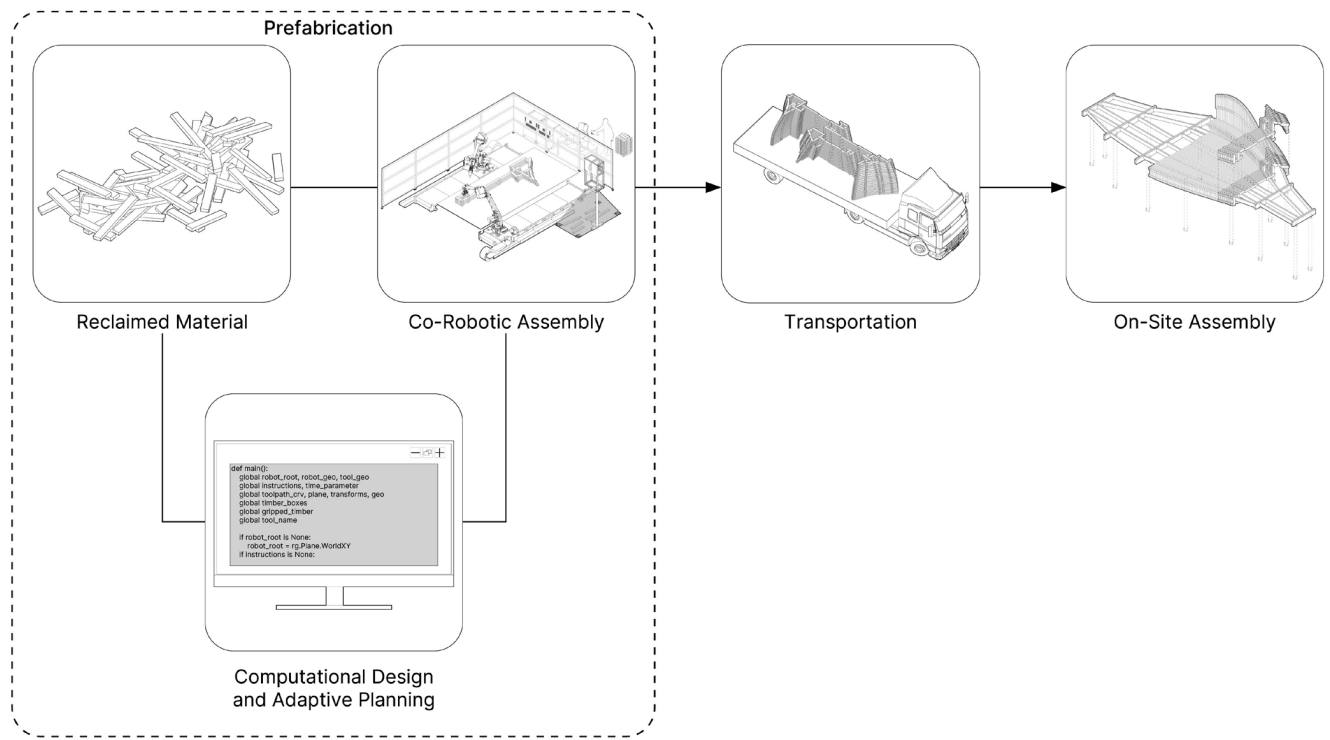

3   Construction process for the case study structure.

are parallel, we can create side-grain lap connections with two fasteners per overlap, alternating diagonals across layers to avoid collisions. Side-grain laps are preferred here because they can develop structural capacity through side-grain bearing (rather than end-grain), accommodate modest dimensional tolerances, and be easily fastened with a nail gun shooting nails perpendicular to the timber elements.

We developed a custom computational design method to implement this constructive system, suitable for modularization, prefabrication, transportation, and on-site assembly, building upon methods introduced in our previous research projects (Ruan and Adel 2023; Adel 2020, 2022; Craney and Adel 2020). This method can be generalized to arbitrary flat surfaces or moderately single or doubly curved surfaces, primarily constrained by the required overlaps between elements in neighboring layers. Fig. 4 illustrates the primary steps of the computational design method for Timbrelyn. Below, we detail each of the primary steps.

Inputs: We implemented the computational design method in Rhinoceros 3D (McNeel n.d.a) and its integrated visual programming environment, Grasshopper (McNeel n.d.b). The computational design method requires a set of input curves (Fig. 4a), which must be explicitly modeled. These curves define the geometric configuration of the installation, which, in the case of Timbrelyn, was informed by the topography of the landscape and the overall program (e.g., a seating area and a stage). In addition, we define a set of parameters and constraints related to the constructive system and fabrication process, such as the timber cross-sectional dimension (1.5 in by 3.5 in for 2×4 lumber), the working envelope of the robotic setup, transportation limits, and a weight constraint for each subassembly. Modification of the inputs retriggers the computation, allowing the digital model to be interactively designed and iteratively refined.

Timber element generation: From the input curves, we construct a set of non-uniform rational B-splines (NURBS) surfaces (Fig. 4b). We then generate a sequence of $N$ parameterized planar contour curves $\mathscr{C} = \{c_i\}_{i=1}^{N}, \; c_i : [0,1] \to \mathbb{R}^3$ by intersecting each surface with a sequence of planes, oriented normal to main longitudinal axis of the installation and spaced equally apart by the element thickness (e.g., 1.5 in. for nominal 2×4 lumber) (Fig. 4c). This cross-sectional slicing approach, rather than longitudinal slicing, yields smaller contours that correspond to smaller subassemblies, which enable better utilization of short timber elements and better fit within the working envelope of our fabrication setup.

For each contour $c_i$ we define a sequence of parameter values $T_i = (t_{i,1}, t_{i,2}, \ldots, t_{i,n_i+1})$ with $0 = t_{i,1} < t_{i,2} < \ldots < t_{i,n_i+1} = 1$, where each parameter $t_{i,j} \in [0,1]$ identifies a point $c_i(t_{i,j})$ along the contour. By adjusting the distribution of these parameter values, either at fixed intervals or through a user-defined function, we control the subdivision density and patterning along each contour. The interval pairs $[t_{i,j}, t_{i,j+1}]$ define $n_i$ linear segments $S_i = \{s_{i,j}\}_{j=1}^{n_i}$, where each segment $s_{i,j} = (c_i(t_{i,j}), c_i(t_{i,j+1}))$ represents the central axis of a potential timber element within the full assembly (Fig. 4d). Aggregating over all contours yields the set of all segments $\mathscr{S} = \{S_i\}_{i=1}^{N}$ in the full assembly. This parametric discretization of the contour curves provides control over element lengths, distribution, and patterning, enabling design variation while facilitating the alignment of the digital model with the reclaimed timber inventory and fabrication constraints.

To control element density and patterning, we apply filtering using binary masks (Fig. 4e). Each segment $s_{i,j}$ is assigned a binary mask value $m_{i,j} \in \{0,1\}$ according to a patterning function $M : S \to \{0,1\}$. Applying the mask to the full assembly results in a subset of retained segments $\mathscr{S}^+ = \{s_{i,j} \in \mathscr{S} \mid M(s_{i,j}) = 1\}$.

We define the masking function $M$ to alternate between neighboring layers to generate patterns that enable overlapping side-grain laps. In addition, for some of the layers, we can modify the mask in response to specific boundary conditions (e.g., seating). The selected line segments are then extended at either ends to create overlapping side-grain laps between connected elements, avoiding seam lines that could introduce structural vulnerabilities. We define a function $G : \mathscr{S}^+ \to \mathscr{G}$ that generates the element from the retained segments (Fig. 4f), where $g_{i,j} \in G_i$ represents the element geometry and its attributes (e.g., cross-section, length, cut planes, and pose) and $G_i \in \mathscr{G}$ is a layer of elements in the full assembly. Fig. 4i visualizes the timber length analysis across the full assembly, which informs data-driven design exploration.

Modularization: To enable off-site co-robotic assembly and on-site handling, the installation must be divided into subassemblies $(\hat{G}_1, \hat{G}_2, \ldots, \hat{G}_k)$. Formally, the set of subassemblies is an ordered partition of $\mathscr{G}$, with $\hat{G}_i \cap \hat{G}_j = \emptyset \;\; \forall i \neq j$ and $\bigcup_{k=1}^{K} \hat{G}_k = \{G_1, G_2, \ldots, G_N\}$. Each subassembly is fabricated layer-by-layer in the robotic workcell and transported to the site for final assembly. Three main constraints govern subassembly feasibility:

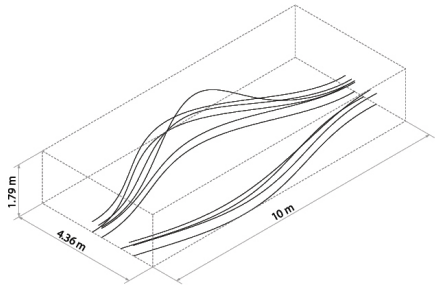
a. Input Curves

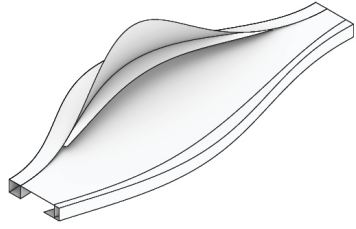
b. NURBS Surface Generation

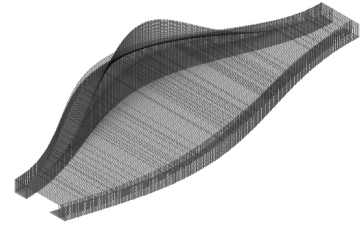
c. Surface Contouring

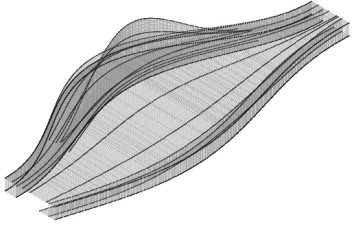
d. Contour Segmentation

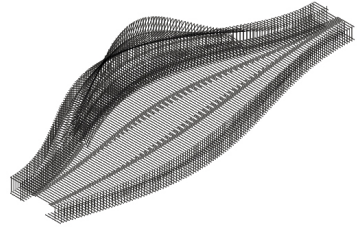
e. Mask-Based Culling

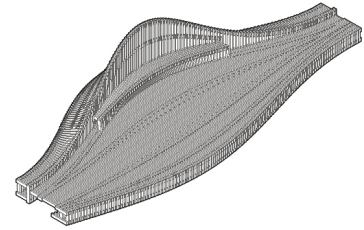
f. Timber Element Generation

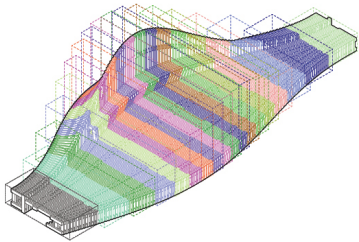
g. Modularization

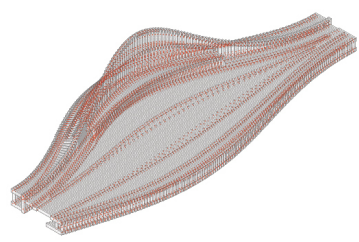
h. Fastening Locations

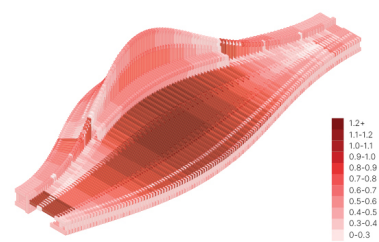
i. Data Visualization

4   Primary steps of the computational design process.

1. Robotic work envelope: The subassembly's bounding box must fit within the cooperative work envelope of the robotic setup (including end effector reach and desired orientation). We approximate this envelope as a box $b_R = (x_R, y_R, z_R) \in \mathbb{R}^3$, corresponding to length, width, and height, respectively.

2. Transportation envelope: Each subassembly must be transported from the prefabrication plant to the project's site. In our case, the transportation truck container's width, length, and height define the transportation envelope, denoted as a box $b_T = (x_T, y_T, z_T) \in \mathbb{R}^3$.

3. Manual-handling weight limit: We impose a maximum weight $w_{max} = 100$ kg per subassembly, so a 4–5 person team can lift a subassembly without powered equipment.

The bounding box of a subassembly is computed as the minimum-volume oriented box that fully encloses all elements within the subassembly, represented as a function $B: \mathscr{G} \to \mathbb{R}^3$. The weight of a subassembly can be computed by assuming a constant density for the element geometry, represented as a function $W: \mathscr{G} \to \mathbb{R}$. A feasible partition of $\mathscr{G}$ satisfies the constraints $W(\hat{G}_k) \leq w_{max}$ and $B(\hat{G}_k) \preceq \min\{b_R, b_T\}$ for all $1 \leq k \leq K$, where $\preceq$ denotes a component-wise (non-strict) inequality. In general, our objective is to minimize the number of subassemblies $K$, minimizing the amount of assembly required on site.

To modularize the full assembly, we employed a two-step approach. In the first step, we employed a greedy algorithm to divide the full assembly by weight. Given the starting index $p = 1$, we find the largest index $q \geq p$ such that $W(\{G_p, G_{p+1}, ..., G_q\}) \leq w_{max}$, output this as a candidate subassembly $\hat{G}_k$, then repeat with $p = q + 1$ until $q = N$. In the second step, we check if each candidate subassembly meets the bounding box constraints; if not, we divide it into smaller subassemblies for fabrication, then join them postfabrication to form the intended subassembly.

By applying this approach, Timbrelyn was partitioned into 18 bespoke subassemblies (Fig. 4g), with seven subassemblies exceeding the cooperative robotic envelope and requiring further partitioning before fabrication.

Generation of fastening connections: In our constructive system, overlapping timber elements are joined through side-grain lap connections fastened with two nails, with the nailing pattern alternating (diagonally) between consecutive layers to prevent potential collisions (Fig. 4h). To determine the fastening locations, the planar overlap regions between elements of adjacent layers are first computed, producing a set of polygons (typically quadrilateral) that define the fastening zones. Each overlap polygon is then offset inwards by a value greater than the minimum distance specified by the National Design Specification for Wood Construction (American Wood Council 2023) to ensure proper fastening and prevent splitting of the timber.

The vertices of the offset polygons are extracted and sorted counterclockwise. For each polygon, we define a consistent starting vertex, computed as its nearest vertex to an arbitrary reference point, to maintain uniform vertex indexing across all overlap polygons. For each layer $i$ of the assembly, a binary mask is applied to the sorted vertices of the overlap polygons to define the positions of the nails. We invert the mask after each layer, such that even and odd indexed layers do not share nail locations. For instance, applying this operation to quadrilateral overlaps results in one diagonal of the quadrilateral defining the location of the two nails in even layers and the other diagonal in the odd layers. This alternation also generalizes to other overlaps, such as three-sided or five-sided polygons and ensures that nails in adjacent layers remain collision-free while maintaining consistent fastening geometry and structural continuity throughout the assembly (Fig. 4.h).[1]

### Adaptive Co-Robotic Assembly

We adapt and advance the co-robotic assembly methods proposed in our previous research (Adel 2022) for the construction of Timbrelyn. Coupled with the developed computational design process, these methods enable an integrated design-to-fabrication framework for the off-site robotic prefabrication of bespoke subassemblies from nonstandard timber inventories. This framework leverages industrial robotic arms for precise assembly, reduces the logistical complexity required for inventory management, and minimizes on-site construction complexity by shifting labor-intensive tasks to off-site prefabrication. The co-robotic assembly method (Fig. 5) forms a closed-loop process that iterates through three computational modules: perception-driven material selection, adaptive processing and assembly, and bidirectional model update, for each element in the subassembly. This loop enables on-the-fly adaptation to reclaimed material variability and as-built assembly tolerances. The technical details of the developed adaptive co-robotic assembly method are beyond the scope of this paper. Below, we provide a high-level overview.

Perception-driven material selection: The assembly method starts with material selection, where the robotic system detects and selects an element to proceed with fabrication

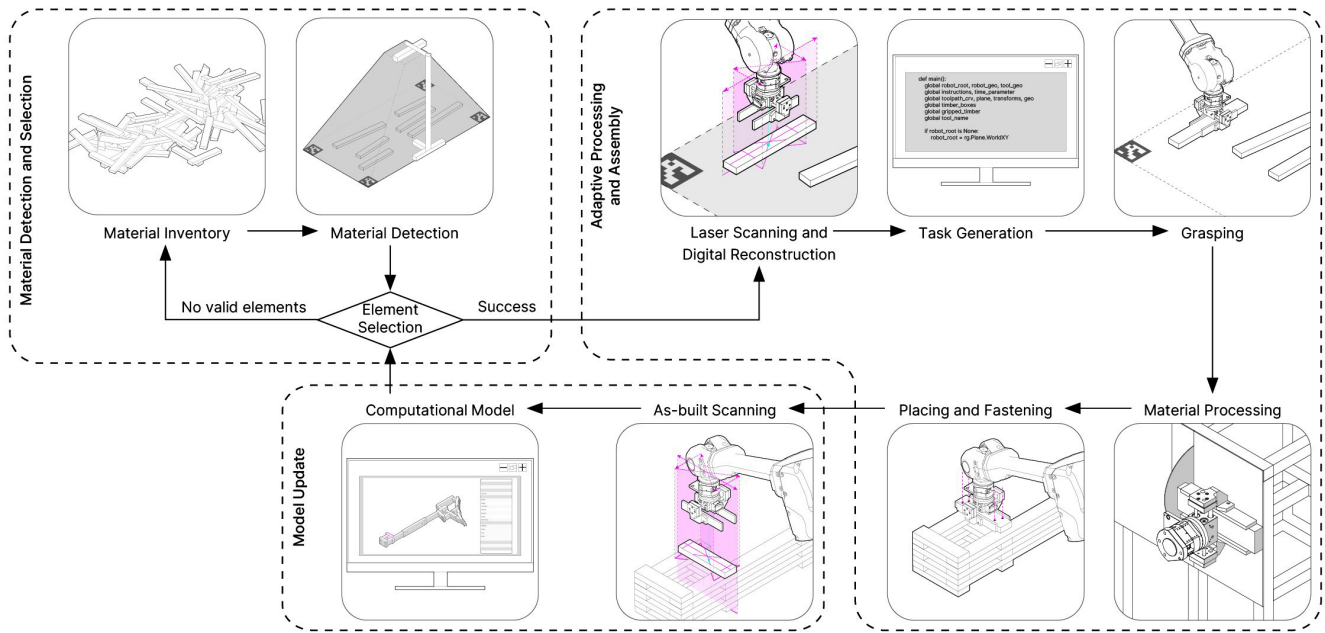

5   Flowchart of the proposed adaptive co-robotic assembly method.

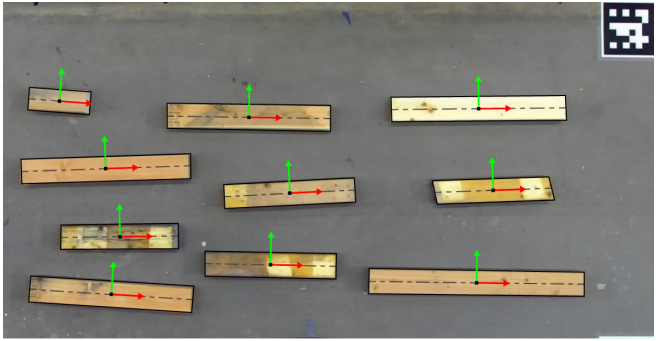

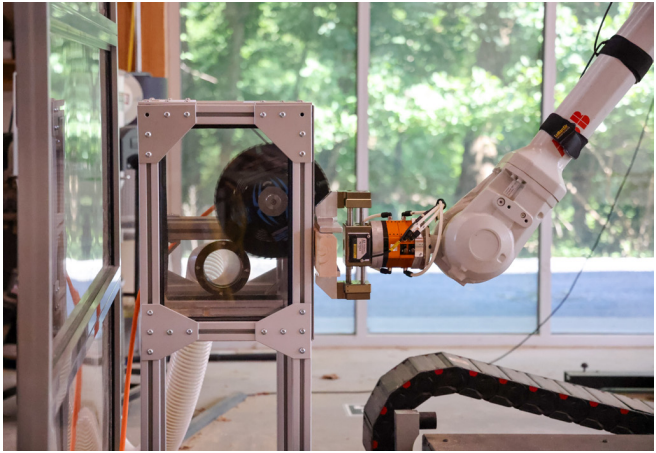

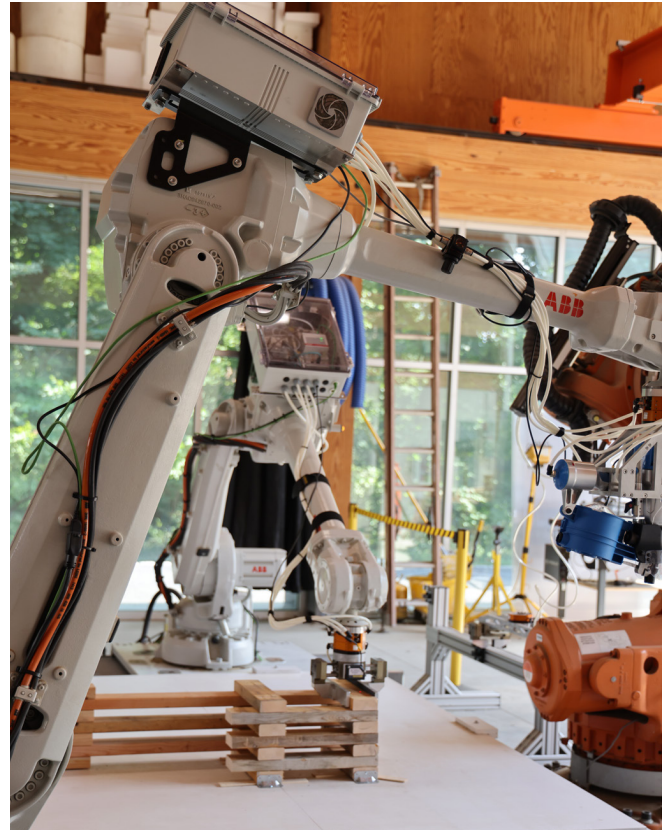

6 Visualization of the perception-driven material detection for adaptive co-robotic assembly.

7 Robotic timber processing on a CNC saw.

8 Robotic timber placement on a subassembly.

based on a set of constraints and user-defined optimality criteria. The element is selected from a subset of the material inventory loaded into a pickup station equipped with an overhead RGB-D (red, green, blue, and depth) camera. The system detects each available element and estimates its two-dimensional (2D) pose (i.e., $(x, y, \theta) \in SE(2)$), contour, and dimensions (length and width). We rectify the image using the method outlined by Štampfl and Ahtik (2023), which utilizes fiducial ArUco markers (Garrido-Jurado, et al. 2014) and projective transformations, mapping image coordinates to the pickup station reference frame to ensure metric accuracy.

Following rectification, we employ the Segment Anything Model (Kirillov et al. 2023) to perform image segmentation to identify discrete objects. We define geometric heuristics, quadrilateral contour shape, aspect ratio, and area thresholds to filter segmentation masks and to remove false positives. The contours are simplified using the Ramer-Douglas-Peucker algorithm (Douglas and Peucker 1973), producing four-point polygons that approximate the boundaries of individual elements (Fig. 6). The dimensions of the elements are computed from their contours and evaluated against the defined geometric constraints and optimality criteria (e.g., cutting length constraints and offcut minimization) to identify the most suitable candidate for fabrication. If no element satisfies the constraints, the system notifies the human operator to add additional elements for evaluation and restarts the perception process.

Adaptive processing and assembly: After material selection, the robotic arm scans the selected element using a one-dimensional (1D) laser displacement sensor mounted on the end effector. We adapt a laser-scanning method from our previous work (Adel et al. 2024), reconfiguring it for use with the 1D laser displacement sensor. This method outputs a three-dimensional (3D) point cloud of the element, which is then used to reconstruct the element geometry using clustering and linear regression techniques. This updated geometry provides a refined estimate of the element pose and its dimensions to enable collision-free grasping and material processing.

The robotic arm then proceeds with grasping the element, cutting it with a computer numerical control (CNC) saw to the target length and miter angles (Fig. 7), and placing the element on the assembly stand (Fig. 8) using fabrication

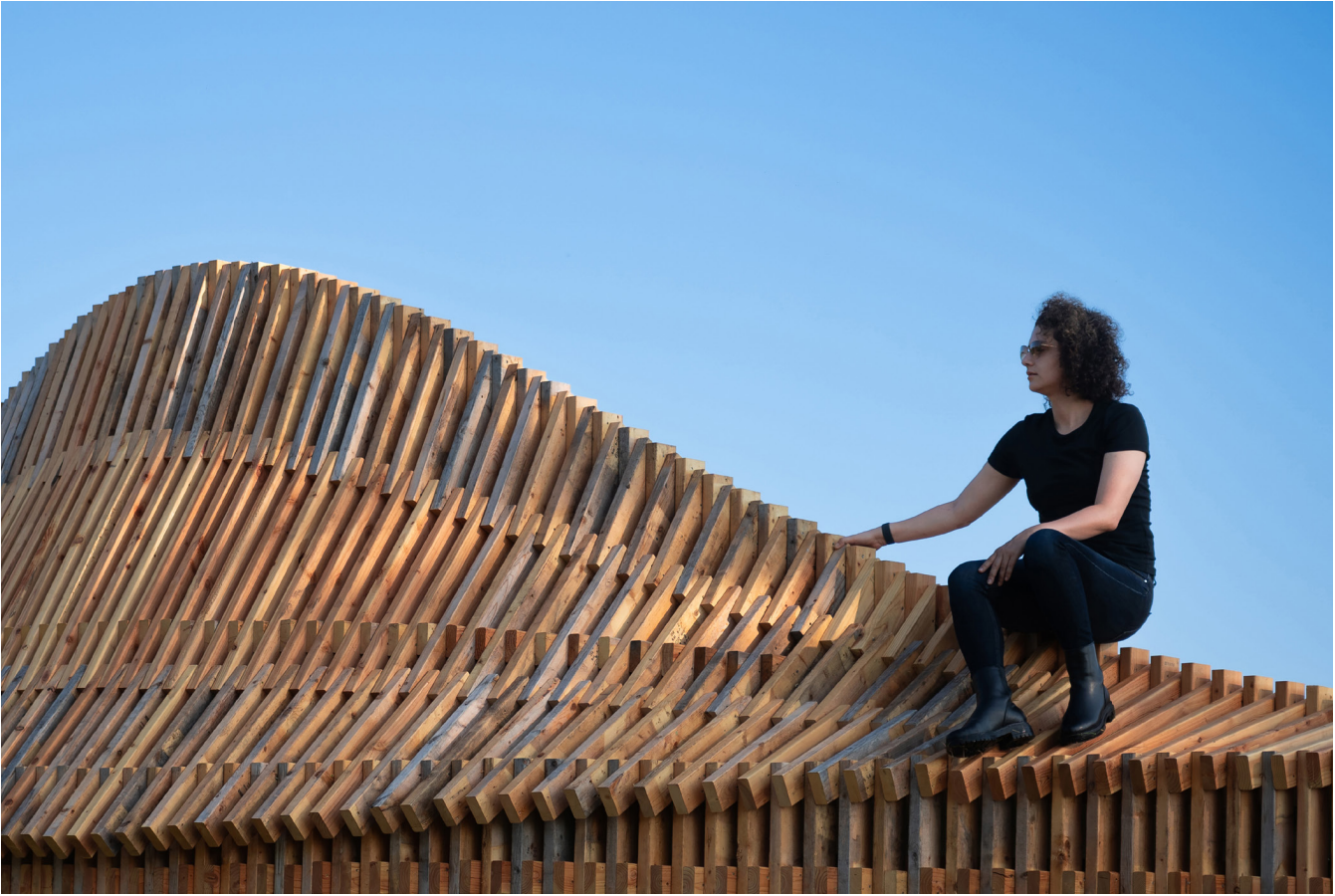

9   Timbrelyn.

parameters derived from the computational model. The arm holds the element in place while a human fabricator nails it to the previously assembled timber elements.

Bidirectional model update: In our current fabrication setup, we do not incorporate any milling or planing operations to address material variability from specified dimensions, which can increase fabrication time and waste. However, this means that height deviations in the subassembly can accumulate as the fabrication progresses, necessitating an adaptive method to avoid either poor contact or excessive force during placement. This is achieved through an update to the target placement pose using the as-built state to compute the proper height for placing each element. After placement and fastening, the robotic arm scans the newly assembled element using the laser displacement sensor, reconstructing its geometry and updating the computational model. This updated computational model informs the assembly of subsequent elements, thereby closing the feedback loop between the computational design model and the physical assembly process.

## RESULTS AND DISCUSSION

The discussed computational design and adaptive co-robotic assembly methods were implemented for the fabrication of Timbrelyn (Fig. 9).

The computational design method established a bidirectional feedback loop between design exploration and quantitative analysis. Parametric design inputs seamlessly generated the geometry and attributes of elements, subassemblies, and their connections. The resulting quantitative metrics, including element counts, length distributions, connection and nail counts; subassembly weights and envelopes; and the ratio of reclaimed to new stock, in turn informed subsequent design decisions. This continuous evaluation accounts for the available material inventory (e.g., reclaimed stock length and quantities) and fabrication, assembly, and transportation constraints, enabling data-driven coupling between architectural intent and site-specific exploration. This process facilitated targeted adjustments such as refining input geometry, porosity, and contour subdivisions to balance material and expressive objectives.

The integrated design-to-fabrication workflow maintained a continuously adaptable computational model of the

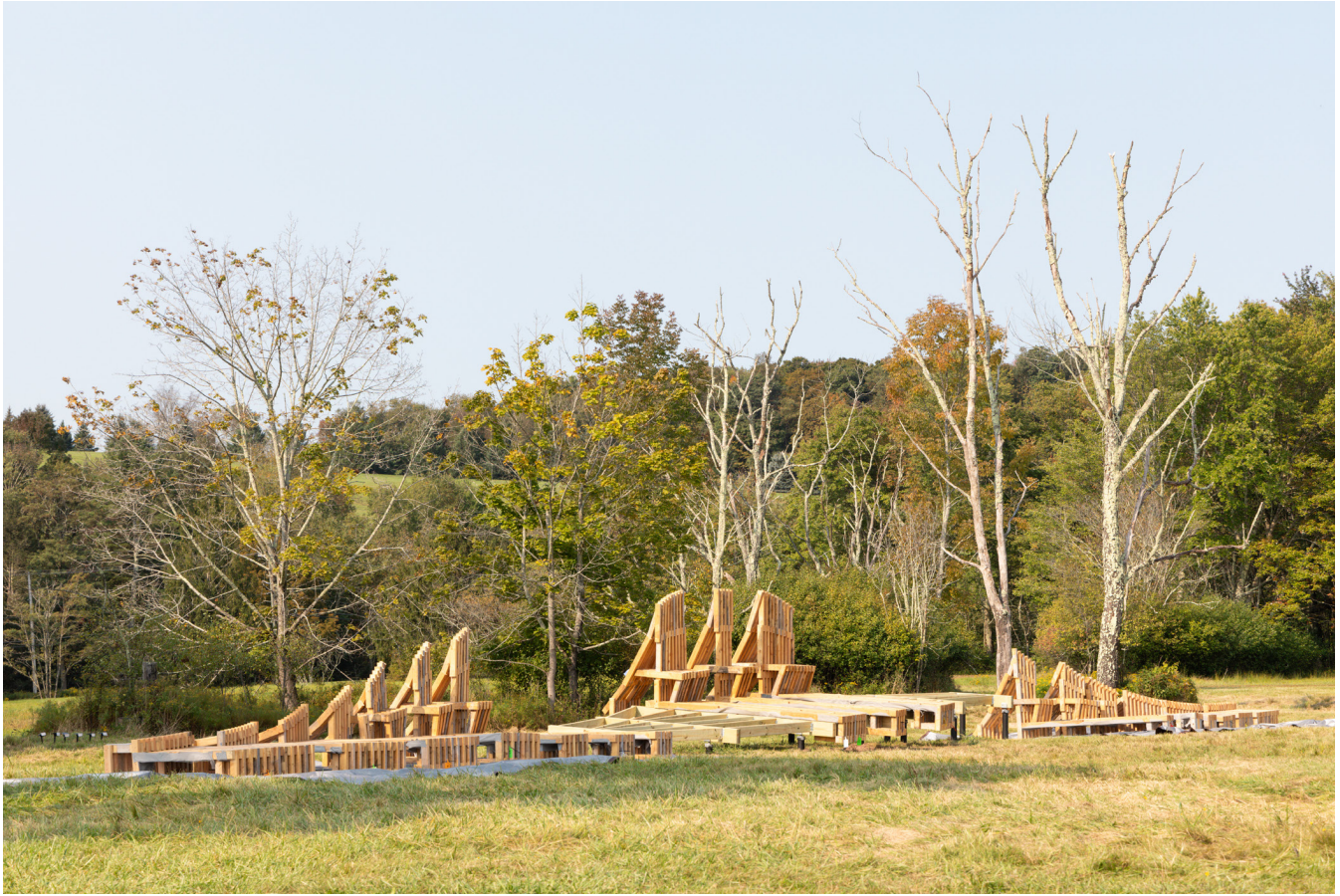

10   Prefabricated subassemblies on the construction site.

design that could be refined up to the start of fabrication. Because geometric definitions, constraints, and fabrication attributes were embedded directly within the computational model, late-stage modifications (e.g., adjusting layer spacing, updating binary masks, or revising subdivision guides) automatically regenerated all associated geometries and fabrication data. This continuity eliminated the rework and translation errors that typically arise from handoffs in conventional phase-based design, planning, and construction approaches. The resulting system established an adaptive, feedback-driven production chain in which each element instance encapsulated the data required for fabrication, including its length, cut planes, and pose.

Employing our developed methods, we fabricated each of Timbrelyn's subassemblies (Fig. 10) using reclaimed timber supplemented by new timber for longer elements where the reclaimed inventory was insufficient. Overall, the installation comprised 1,838 timber elements, of which 1,675 (91.1%) were robotically cut and placed. The remaining 163 elements were too short (under 350 mm in length) to be cut safely by the robot on the CNC saw; however, they were still robotically placed after being cut manually and detected using our material detection method. These short elements were required to fill in structural gaps in the design, and their fabrication follows the same robotic fabrication method as other elements, except the system automatically skips cutting during material processing. This approach allowed us to leverage the precision of robotic assembly throughout the entire fabrication process with minimal impact on the sequence of operations.

Beyond its technical contributions, Timbrelyn demonstrates how adaptive computational design-to-fabrication for material reuse can inform and enhance architectural expression. The hybrid material palette of both newly sourced and weathered, reclaimed timber produced a layered texture that remains visibly legible in the completed structure (Fig. 11). The alternating patterns of timber elements introduced tactile variation and unanticipated visual effects, foregrounding the heterogeneity of reclaimed material as an expressive quality rather than a constraint. In this way, Timbrelyn illustrates how adaptive co-robotic assembly methods and material reuse can serve not only technical and environmental objectives but also contribute to novel experiential and aesthetic dimensions in architecture.

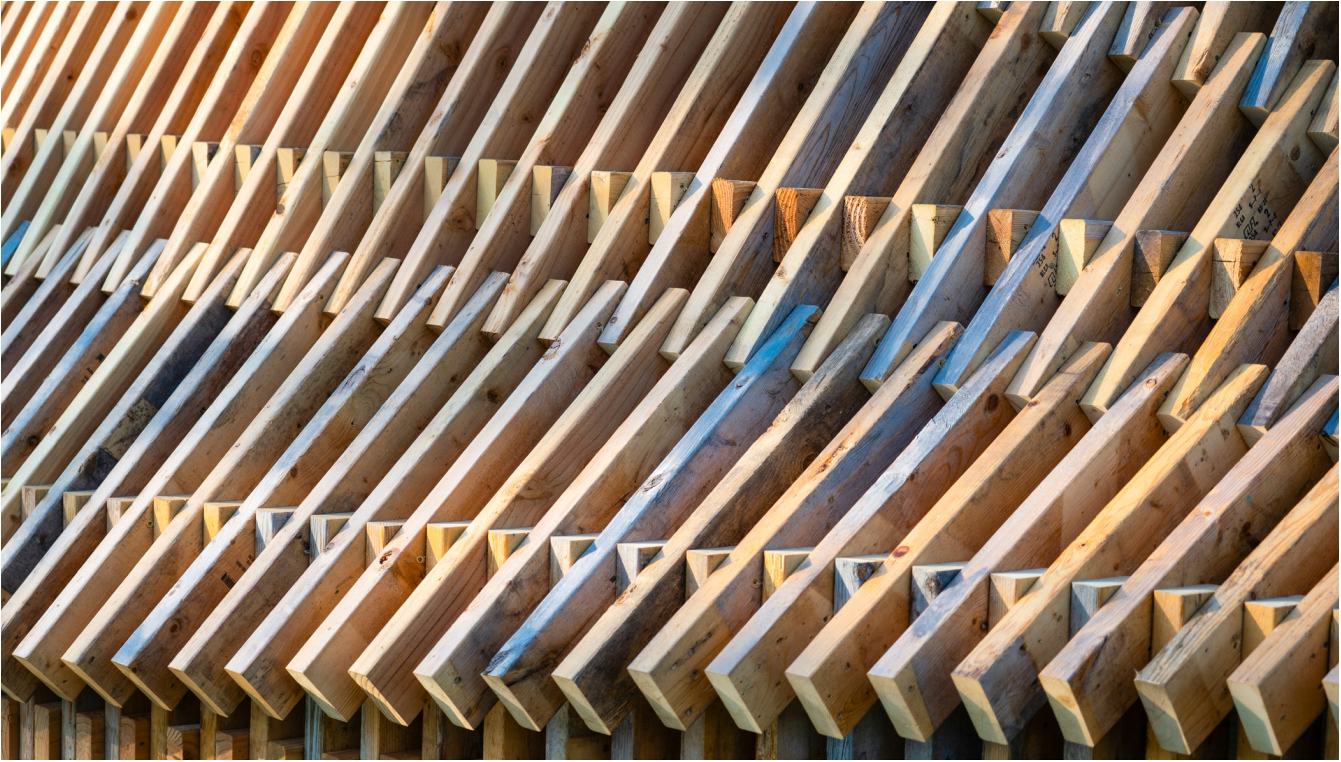

11  Close-up of the visual textural effect of the newly sourced and weathered, reclaimed timber elements.

## CONCLUSION

This paper presented a fully integrated design-to-fabrication framework that combines data-driven computational design with adaptive co-robotic assembly for material reuse with reclaimed timber. The framework was validated through the design and construction of Timbrelyn, a built case-study installation. On the design side, the computational design method translated architectural intent into fabrication-ready, data-rich models that incorporated robotic fabrication constraints, modularization requirements for transport and handling, and reclaimed material stock augmented with new timber when necessary. On the fabrication side, the integration of perception-driven material selection, adaptive processing and assembly, and bidirectional model updating established a closed feedback loop between digital and physical processes. Together, these methods constitute a data-driven, material-aware, and adaptive framework that supports the reuse of heterogeneous timber stock while minimizing offcut waste. This research demonstrates the potential of circular, feedback-driven construction automation to enable the reuse of reclaimed materials while generating new forms of architectural expression.

### Future Work

While the results validate the potential of the proposed framework, several limitations remain that motivate ongoing and future research. The computational design method could be extended by formalizing and representing the connectivity of the timber structure as a graph (Adel 2020; Adel et al. 2018), in which nodes represent elements and edges represent their connections. Such a representation would enable explicit reasoning about global connectivity, seam avoidance, and constructability, while directly linking to structural analysis and optimization. Building on this foundation, the current layer-based constructive system can be extended to frame or spatial assemblies that incorporate both side-grain and end-grain lap joints.

In addition, the current perception system is limited to geometric detection of material features (e.g., length, width, pose). Future extensions could integrate visual and multimodal sensing to identify material defects (e.g., knots, cracks, holes) and contaminants (e.g., metal fasteners, connectors) that introduce processing or safety risks. Incorporating these material attributes into the computational workflow would enable more informed material selection and allocation, advancing the integration of perception, optimization, and adaptive fabrication for the reuse of reclaimed timber and other reclaimed materials.

## ACKNOWLEDGMENTS

The authors gratefully acknowledge and thank all the people who directly or indirectly contributed to the research presented in this paper, as well as to the case-study built installation, Timbrelyn. The full project credits for Timbrelyn are listed below:

Principal Investigator and Project Lead: Arash Adel, Adel Research Group (ARG)

Research, Design, and Fabrication Assistants: Daniel Ruan, Ruxin Xie, Zhengyi Chen

Student Assistants: Zoe Cheung, Zhuofan Ma

On-site Assembly Assistants: Rimervi Mendez Vasquez, Carlos Lantigua, Autumn Siedlik

Structural Engineers: TYLin (Justin Den Herder, Amy Harrington)

Diagrams: Ruxin Xie

Photographers and Videographers: Arash Adel, Daniel Ruan, Ruxin Xie, Thanut Sakdanaraseth

Video Editors: Princeton Broadcast Center (Katie E. Weinstein, Kayce West, Jared Montano, Eric Alonzo)

Supported by Princeton University School of Architecture, Bethel Woods Center for Arts

Special Thanks: Monica Ponce de Leon, Paul Lewis, Salma Mozaffari, Neal Hitch, Neal Lucas Hitch

### NOTES

1. We developed an automated nailing process with wooden nails using a second robotic arm. However, the timber elements were splitting due to the force of the wooden nail and the cross-sectional dimension of the elements. Due to time constraints, we decided to integrate manually nailing of metal nails into the process (using a nail gun) instead of automated nailing of wooden nails with the robotic arm.

## IMAGE CREDITS
Figures 1, 9, 11: Thanut Sakdanaraseth
Figure 8: Zoe Cheung
All other drawings and images by the authors.

---

**Arash Adel** is an Assistant Professor of Architecture and Associated Faculty of Computer Science at Princeton University, where he directs the Adel Research Group (ARG). ARG is an interdisciplinary laboratory that conducts research at the intersection of robotics, artificial intelligence, and computational design with the overarching goal of advancing resource-aware and labor-conscious architecture and construction practices.

Prior to joining Princeton University in 2023, Adel was an Assistant Professor of Architecture at the University of Michigan. Adel received his Doctorate in Architecture from ETH Zurich and his Master's in Architecture from Harvard University.

---

**Daniel Ruan** is a research associate at ARG. He is currently a PhD in student at Princeton University's School of Architecture. He graduated from Rensselaer Polytechnic Institute with his Bachelor of Architecture and Master of Science in Architectural Science, followed by the Master of Science in Digital and Material Technologies program at the University of Michigan. His interests include the interface between humans and digital processes, in how the collaborative efforts of both can heighten the design process of architecture in research, teaching, and practice.

---

**Ruxin Xie** is a research associate at ARG. She holds a Master of Architecture and a Master of Science in Digital and Material Technologies from the University of Michigan. Prior to her current role, Ruxin worked as a technical designer at Gensler in San Jose, California. She is dedicated to employing computational design methods and cutting-edge digital fabrication technologies to advance the architecture and construction disciplines.